\documentclass[letterpaper]{article}

\usepackage{amsthm}
\usepackage{times}
\usepackage{helvet}
\usepackage{courier}
\usepackage{url}
\usepackage[pdftex]{graphicx}
\usepackage[utf8]{inputenc}
\usepackage[T1]{fontenc}
\usepackage{booktabs}
\usepackage{amsfonts}
\usepackage{nicefrac}
\usepackage{microtype}
\usepackage{tikz}
\usepackage{grffile}
\usetikzlibrary{shapes.misc, positioning}
\usepackage{thmtools,thm-restate}
\usepackage{parskip}
\usepackage{authblk}
\usepackage{tcolorbox}
\usepackage{wrapfig}
\usepackage{amsmath}
\usepackage{amssymb}
\usepackage{mathtools}
\usepackage[numbers]{natbib}
\usepackage{multirow}
\usepackage{color}
\usepackage{hyperref}
\usepackage{colortbl}
\usepackage{algpseudocode}
\usepackage[none]{hyphenat}
\usepackage{floatrow}
\usepackage{macros}
\usepackage[pro]{fontawesome5}
\usepackage{basic}

\usepackage{microtype}
\usepackage{enumitem}
\usepackage{xcolor}
\usepackage[normalem]{ulem}
\useunder{\uline}{\ul}{}
\usepackage{tcolorbox}   
\usepackage{subfig}
\usepackage{listings}
\tcbuselibrary{breakable}
\usepackage{hyperref}

\hypersetup{
    colorlinks=true,
    linkcolor=purple,
    urlcolor=teal,
    linktoc=all,
    citecolor=teal
}

\newtcolorbox{boxI}[2][]{
  colback   = blue!5,
  colframe  = black!40,
  fonttitle = \bfseries,
  title     = {#2},
  sharp corners,
  boxrule   = .4pt,
  breakable,
  #1
}

\title{Time To Impeach LLM-as-a-Judge: Programs are the Future of Evaluation}

\author[$\dagger$]{Tzu-Heng~Huang}
\author[$\dagger$]{Harit~Vishwakarma}
\author[$\dagger$]{Frederic~Sala}

\affil[$\dagger$]{University of Wisconsin-Madison}
\affil[ ]{\footnotesize{\texttt{\{thuang273, hvishwakarma, fredsala\}@wisc.edu}}}

\begin{document}

\maketitle

\begin{abstract}
Large language models (LLMs) are widely used to evaluate the quality of LLM generations and responses, but this leads to significant challenges: high API costs, uncertain reliability, inflexible pipelines, and inherent biases. 
To address these, we introduce \textbf{PAJAMA} (\underline{P}rogram-\underline{A}s-a-\underline{J}udge 
 for 
\underline{A}utomated \underline{M}odel \underline{A}ssessment), a new alternative that uses LLMs to \textbf{\emph{synthesize executable judging programs}} instead of directly scoring responses.
These synthesized programs can be stored and run locally, costing orders of magnitude less while providing interpretable, and auditable judging logic that can be easily adapted.
Program-based judges mitigate biases, improving judgment consistency by \textbf{15.83\%} and reducing biased responses by \textbf{23.7\%} on average compared to a Qwen2.5-14B-based LLM-as-a-judge. 
When program judgments are distilled into a model, PAJAMA outperforms LLM-as-a-judge on the challenging CHAT-HARD subset of RewardBench, outperforming metrics by \textbf{2.19\%} on Prometheus and \textbf{8.67\%} on the JudgeLM dataset, all at three orders of magnitude lower cost.
\end{abstract}

\section{Introduction}

%
General-purpose large language models (LLMs) are now the standard for \emph{automated evaluation} of generative model responses.
In the typical LLM-as-a-judge setup, a judge LLM evaluates a user query with two LLM-generated answers, then selects the better one~\cite{wei2024systematic}.
These judgments can be viewed as preference labels that enable ranking LLMs by performance~\cite{chiang2024chatbot, zheng2023judging}, efficiently verifying LLM agent behaviors~\cite{zhuge2024agent}, curating datasets~\cite{wettig2024qurating}, or distilling into specialized reward models~\cite{christiano2017deep}.

%
Despite their promise, current LLM-as-a-judge approaches have several drawbacks:
\begin{itemize}[topsep=0pt, itemsep=0pt, partopsep=0pt, parsep=0pt, leftmargin=*]
    \item \textbf{High cost}: 
    Querying state-of-the-art models, e.g., Open AI o3, or Claude 3 Opus~\cite{TheC3}, can run into thousands of dollars for large evaluation sets. 
    For example, researchers spent about \$4,000 to generate 100K high-quality judgments using GPT-4~\cite{zhu2023judgelm}.
    %
    %
    \item \textbf{Uncertain reliability}:
    Although LLMs can explain their verdicts, their outputs may exhibit inconsistent reliability in adhering to rubrics~\cite{yu2024xfinder}.
    \item \textbf{Inflexible pipelines}: 
    Any minor change to the evaluation rubric requires re‑running the entire pipeline, incurring additional expenses.
    \item \textbf{Inherent biases}: 
    LLMs trained on web datasets encode social and stylistic biases (e.g., gender preferences, or favoring emojis), which can skew decisions~\cite{adila2024discovering, chen-etal-2024-humans, ye2024justice}.
\end{itemize}
%

%
We address these by relying on a \emph{simple but powerful} notion.
Instead of prompting an LLM for preferred answers, \emph{\textbf{we ask it to generate the judging logic it would use and encode it into an executable program.}}
In other words, the model is asked to synthesize a compact function, e.g., a few lines of Python, that encodes its evaluation criteria.
These are executed to obtain a quality score for each response.

This \emph{program synthesis}-style strategy offers several advantages. 
\textbf{\emph{First}}, API costs now scale with the number of generated programs, not the dataset size, significantly reducing expenses. 
Once generated, judging programs can be stored and executed locally for any new query at no additional cost. 
\textbf{\emph{Second}}, synthesized programs are interpretable, allowing practitioners to audit each line, refine rubrics, or insert heuristic rules to minimize bias.
\textbf{\emph{Third}}, by exposing the full decision logic, this approach turns a black-box judging model into a transparent, inspectable evaluation process.

%
While promising, translating LLM judgments into executable code raises new challenges. 
It is common for synthesized programs to be repetitive, reusing similar criteria with minor variations. 
To encourage diversity, we introduce six distinct criteria---\emph{\textbf{each expressible as code}}---to guide LLMs in generating useful programs. 
Additionally, individual program outputs can be noisy, and different programs may capture complementary signals. 
We address this by combining our approach with the weak supervision framework~\cite{ratner2016data, ratner2017snorkel, ratner2019training, fu2020fast}.
By modeling program outputs, we aggregate program judges into a collective signal that can outperform naïve ensembling approaches like majority voting.

%
\noindent \textbf{Results and Contributions.} To tackle the above, we propose \textbf{PAJAMA} (\underline{P}rogram-\underline{A}s-a-\underline{J}udge \underline{A}utomated \underline{M}odel \underline{A}ssessment), a lower-cost and lower-bias evaluation system that relies on synthesized judging programs. 
Each program can serve directly as a judge or have its judgments distilled into a reward model for improved generalization. 
For example, when distilled into a reward model, PAJAMA outperforms LLM‑as‑a-judge‑distilled reward models on the challenging CHAT-HARD subset of RewardBench~\cite{lambert2024rewardbench}, achieving \textbf{+2.19\%} on Prometheus~\cite{kim2023prometheus} and \textbf{+8.67\%} on JudgeLM dataset~\cite{zhu2023judgelm}, while decreasing API costs by approximately 3500$\times$ and 2500$\times$, respectively.
We also test PAJAMA on high-bias responses, observing that it produces consistent correct responses and improved error rate over LLM-as-a-judge. 
Across four common bias types, it enhances consistency by \textbf{15.83\%} and reduces the biased‑answer win rate by \textbf{23.7\%} relative to Qwen2.5‑14B. 

\section{Evaluation System: PAJAMA}

We start with an overview of PAJAMA's general workflow, followed by the problem setup (Sec.~\ref{sec:setup}), then describe program synthesis for judging code (Sec.~\ref{sec:program_judges}) and discuss how we combine multiple judgments to produce an aggregated evaluation decision (Sec.~\ref{sec:weak_supervision}).

\begin{figure}[t!]
    \begin{center}
    \includegraphics[width=0.8\linewidth]{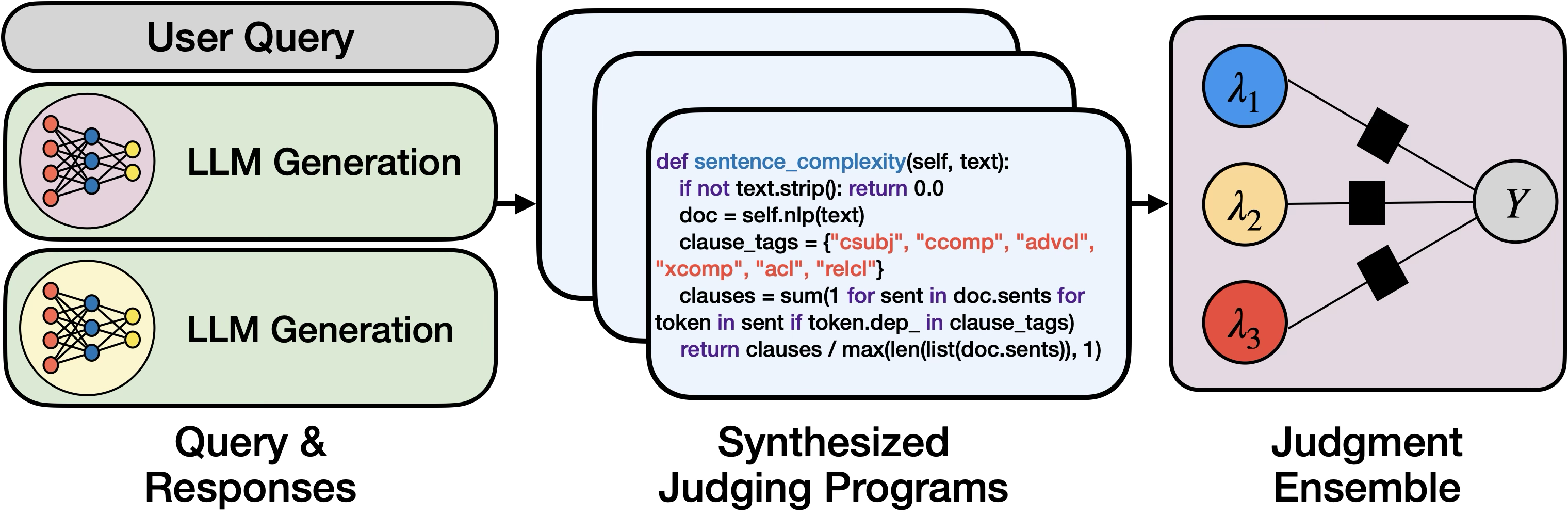}
    \end{center}
    \centering
    \caption{PAJAMA's General Workflow.}
    \label{fig:framework}
\end{figure}

\noindent \textbf{General Workflow.} 
The workflow of PAJAMA is illustrated in Fig.~\ref{fig:framework}.
%
%
%
%
%
First, users collect queries and LLM generation pairs, then create prompts to guide LLMs in generating programs with judging logic.
Prompts can vary in their design and specified evaluation criteria, enabling the creation of multiple judging programs. 
Using weak supervision, aggregated preference labels are inferred from program outputs.
Finally (and optionally), these judgments can be used to train a distilled reward model for local use.

\subsection{Problem Setup}
\label{sec:setup}
We consider user queries and responses drawn from the space of free-form text, $\Sigma^\star$, e.g., all possible natural language strings. 
We denote the space of queries as $\mathcal{X}$ and the space of responses as $\mathcal{Y}$.
For any query $x \in \mathcal{X}$, we evaluate two responses, $y_1, y_2 \in \mathcal{Y}$, generated by the same or different LLMs.
%
%
%
Other approaches directly score each candidate generation, i.e., assign a score to each $y$.

%
%
%
Ground-truth assessments for reward or evaluation models are accurate but costly and slow due to human annotations.
Using LLM-as-a-judge is faster but incurs high inference costs, and may embed LLM biases into ground truth and downstream models.
We address these issues with synthesized judging programs as an alternative.

\subsection{Program Synthesis}
\label{sec:program_judges}
%
%
%
We propose a prompting template to query LLMs for synthesizing programmatic judges.
Using GPT-4o~\cite{hurst2024gpt}, we generate these judges as follows:
\begin{boxI}{Prompt used to synthesize judging programs.}
You are a judge tasked with evaluating LLM-generated responses to a given question. Write your evaluation logic as Python code, returning a numeric score for a response where higher values indicate better quality. Use third-party libraries (e.g., embedding models, nlp metrics) as needed.

\textbf{def judging\_function(query, response)}:
\end{boxI}
We chose this approach for its simplicity, but we note that \emph{more sophisticated approaches to program synthesis can be seamlessly swapped in}. 
Next, we propose six distinct criteria that can be incorporated into the program synthesis prompt and thus be encoded into executable Python code.
%
We describe each criterion below. We likewise note that these can be easily swapped for other criteria relevant to the particular task of interest. 
\begin{itemize}[topsep=0pt, itemsep=0pt, partopsep=0pt, parsep=0pt, leftmargin=*]
    \item \textbf{Structure}:
    A judge evaluates text by analyzing features such as transition markers (e.g., ``therefore,'' ``however''), sentence count, paragraph length, and the presence of headings, questions, or emphasized text. 
    More markers and structured elements indicate better quality.
    \item \textbf{Relevance}:
    A judge assesses semantic alignment between the question and response.
    For example, one approach uses TF-IDF to compute cosine similarity for lexical overlap. 
    Another can employ semantic embeddings to measure deeper contextual similarity~\cite{multi2024m3}.
    \item \textbf{Readability}:
    A judge analyzes grammar errors, information density, and counts repetitive words.
    It also includes third-party libraries to compute readability metrics like the Flesch--Kincaid grade level~\cite{kincaid1975derivation}.
    \item \textbf{Bias}: 
    A judge evaluates response objectivity using sentiment analysis and regex patterns to detect stance and biased keywords, ensuring neutrality.
    \item \textbf{Factuality}:
    A judge assesses factual accuracy, using fine-tuned BERT models to verify content correctness~\cite{feng-etal-2023-factkb}.
    \item \textbf{Safety}:
    A judge employs fine-tuned BERT models, trained to detect hate speech or harmful content, to ensure responses are safe and appropriate~\cite{vidgen2021lftw}.
\end{itemize}

We note that these judging principles,  evaluation rubrics, and keyword lists are all generated by the GPT4o. 
Each of the resulting programs can function as an independent judge to assess the quality of LLM responses.

\subsection{Judgment Ensemble}
\label{sec:weak_supervision}

The outputs of program-based judges can be noisy or incomplete. For this reason, we seek to \emph{combine them}. Doing so enables reducing noise and taking advantage of complementary signals. 
We perform the aggregation by borrowing weak supervision techniques. 

Suppose we generate $m$ judging programs, each implementing a scoring function $\lambda_i : \mathcal{X} \times \mathcal{Y} \to [0,1]$.
For a prompt $x$ and two candidate responses $y_1$ and $y_2$, we discretize their scores by defining
\[
\bar{\lambda}_i(x, y_1, y_2) =
\begin{cases}
+1, & \text{if } \lambda_i(x, y_1) > \lambda_i(x, y_2), \\
-1, & \text{otherwise}.
\end{cases}
\]
In words, a judge outputs $+1$ when it prefers $y_1$ over $y_2$, and $-1$ when it prefers the opposite.

Borrowing the terminology of weak supervision \cite{ratner2019training, shin2021universalizing}, we model a joint distribution over the noisy outputs $\bar{\lambda}_i$, conditioned on the (latent) true preference label $Y(x,y_1,y_2) \in \{+1,-1\}$, as
\begin{align*}
\Pr \Big ( \bar{\lambda}_1, \ldots, \bar{\lambda}_m \mid x,y_1,y_2 \Big )  = \frac{1}{Z_\theta}\exp \Big( -\theta_i \sum_{i=1}^m \bar{\lambda}_i Y \Big),
\end{align*}
where $\theta_i$ denotes the reliability weight, i.e., accuracy, learned for judge~$i$, and $Z_\theta$ is the normalizing partition function ensuring a valid probability distribution.
Once these weights are learned, the ensemble can infer a consensus preference label for any triple $(x, y_1, y_2)$~\cite{ratner2016data, ratner2017snorkel, ratner2019training, fu2020fast}.
These inferred preference labels can be used to evaluate LLM generations directly, or be distilled into a reward model~\cite{christiano2017deep}. The distilled model is often able to generalize beyond the constituent programs \cite{Shin25}.

\section{Experiments}

\begin{table*}[t!]
\centering
\caption{PAJAMA achieves competitive preference label accuracy (\%) at a significantly lower cost than LLM-as-a-judge.}
\label{tab:full_table}
\resizebox{\linewidth}{!}{%
\begin{tabular}{@{}l|c|cc|cc|cccccc@{}}
\toprule
 &  & \multicolumn{2}{c|}{} & \multicolumn{2}{c|}{} & \multicolumn{6}{c}{\textbf{\begin{tabular}[c]{@{}c@{}}RewardBench \\ (Out-of-Domain)\end{tabular}}} \\ \cmidrule(l){7-12} 
 &  & \multicolumn{2}{c|}{\multirow{-2}{*}{\textbf{\begin{tabular}[c]{@{}c@{}}Estimated Cost\\ (using GPT4o)\end{tabular}}}} & \multicolumn{2}{c|}{\multirow{-2}{*}{\textbf{\begin{tabular}[c]{@{}c@{}}Evaluation Set\\ (In-Domain)\end{tabular}}}} & \multicolumn{2}{c|}{Chat} & \multicolumn{2}{c|}{Chat Hard} & \multicolumn{2}{c}{Reasoning} \\ \cmidrule(l){3-12} 
\multirow{-3}{*}{} & \multirow{-3}{*}{\begin{tabular}[c]{@{}c@{}}Dataset \\ Size\end{tabular}} & \multicolumn{1}{c|}{LLM-as-a-Judge} & \cellcolor[HTML]{ECF4FF}\textbf{PAJAMA} & \multicolumn{1}{c|}{LLM-as-a-Judge} & \cellcolor[HTML]{ECF4FF}\textbf{PAJAMA} & \multicolumn{1}{c|}{LLM-as-a-Judge} & \multicolumn{1}{c|}{\cellcolor[HTML]{ECF4FF}\textbf{PAJAMA}} & \multicolumn{1}{c|}{LLM-as-a-Judge} & \multicolumn{1}{c|}{\cellcolor[HTML]{ECF4FF}\textbf{PAJAMA}} & \multicolumn{1}{c|}{LLM-as-a-Judge} & \cellcolor[HTML]{ECF4FF}\textbf{PAJAMA} \\ \toprule
\rowcolor[HTML]{EFEFEF} 
Prometheus & 59,928 & \multicolumn{1}{c|}{\cellcolor[HTML]{EFEFEF}\$183.67} & \cellcolor[HTML]{ECF4FF}\$0.053 & \multicolumn{1}{c|}{\cellcolor[HTML]{EFEFEF}---} & \cellcolor[HTML]{ECF4FF}--- & \multicolumn{1}{c|}{\cellcolor[HTML]{EFEFEF}91.90} & \multicolumn{1}{c|}{\cellcolor[HTML]{ECF4FF}75.00} & \multicolumn{1}{c|}{\cellcolor[HTML]{EFEFEF}38.38} & \multicolumn{1}{c|}{\cellcolor[HTML]{ECF4FF}\textbf{40.57}} & \multicolumn{1}{c|}{\cellcolor[HTML]{EFEFEF}82.26} & \cellcolor[HTML]{ECF4FF}59.54 \\
JudgeLM & 55,751 & \multicolumn{1}{c|}{\$133.04} & \cellcolor[HTML]{ECF4FF}\$0.053 & \multicolumn{1}{c|}{82.88} & \cellcolor[HTML]{ECF4FF}72.82 & \multicolumn{1}{c|}{95.25} & \multicolumn{1}{c|}{\cellcolor[HTML]{ECF4FF}66.76} & \multicolumn{1}{c|}{48.68} & \multicolumn{1}{c|}{\cellcolor[HTML]{ECF4FF}\textbf{57.35}} & \multicolumn{1}{c|}{73.71} & \cellcolor[HTML]{ECF4FF}45.71 \\
\rowcolor[HTML]{EFEFEF} 
PandaLM & 233,227 & \multicolumn{1}{c|}{\cellcolor[HTML]{EFEFEF}\$300.37} & \cellcolor[HTML]{ECF4FF}\$0.053 & \multicolumn{1}{c|}{\cellcolor[HTML]{EFEFEF}74.47} & \cellcolor[HTML]{ECF4FF}63.26 & \multicolumn{1}{c|}{\cellcolor[HTML]{EFEFEF}94.41} & \multicolumn{1}{c|}{\cellcolor[HTML]{ECF4FF}83.24} & \multicolumn{1}{c|}{\cellcolor[HTML]{EFEFEF}42.43} & \multicolumn{1}{c|}{\cellcolor[HTML]{ECF4FF}31.91} & \multicolumn{1}{c|}{\cellcolor[HTML]{EFEFEF}79.60} & \cellcolor[HTML]{ECF4FF}58.52 \\ \bottomrule
\end{tabular}%
}
\end{table*}

\begin{table*}[t!]
\centering
\caption{PAJAMA is more bias-resistent over LLM-as-a-judge.}
\label{tab:bias_reduction}
\resizebox{\linewidth}{!}{%
\begin{tabular}{@{}l|c|cc|cc|cc|cc@{}}
\toprule
 & \textbf{Position} & \multicolumn{2}{c|}{\textbf{Gender}} & \multicolumn{2}{c|}{\textbf{Rich-content}} & \multicolumn{2}{c|}{\textbf{Reference}} & \multicolumn{2}{c}{\textbf{Average}} \\ \cmidrule(l){2-10} 
 & Consistency & \multicolumn{1}{c|}{Consistency} & \begin{tabular}[c]{@{}c@{}}Biased Response \\ Win Rate\end{tabular} & \multicolumn{1}{c|}{Consistency} & \begin{tabular}[c]{@{}c@{}}Biased Response \\ Win Rate\end{tabular} & \multicolumn{1}{c|}{Consistency} & \begin{tabular}[c]{@{}c@{}}Biased Response \\ Win Rate\end{tabular} & \multicolumn{1}{c|}{\begin{tabular}[c]{@{}c@{}}Consistency\\ (4 biases)\end{tabular}} & \begin{tabular}[c]{@{}c@{}}Biased Response \\ Win Rate (3 biases)\end{tabular} \\ \toprule
\rowcolor[HTML]{EFEFEF} 
Llama3-8B & 45.07 & \multicolumn{1}{c|}{\cellcolor[HTML]{EFEFEF}{\ul 50.59}} & {\ul 11.27} & \multicolumn{1}{c|}{\cellcolor[HTML]{EFEFEF}{\ul 60.33}} & 45.54 & \multicolumn{1}{c|}{\cellcolor[HTML]{EFEFEF}53.05} & 53.52 & \multicolumn{1}{c|}{\cellcolor[HTML]{EFEFEF}{\ul 52.41}} & 36.78 \\
Qwen2.5-7B & 26.76 & \multicolumn{1}{c|}{22.54} & 19.29 & \multicolumn{1}{c|}{\textbf{60.56}} & 44.84 & \multicolumn{1}{c|}{\textbf{60.33}} & 79.81 & \multicolumn{1}{c|}{42.55} & 47.98 \\
\rowcolor[HTML]{EFEFEF} 
Qwen3-8B & {\ul 58.45} & \multicolumn{1}{c|}{\cellcolor[HTML]{EFEFEF}39.20} & \textbf{4.93} & \multicolumn{1}{c|}{\cellcolor[HTML]{EFEFEF}50.70} & \textbf{25.59} & \multicolumn{1}{c|}{\cellcolor[HTML]{EFEFEF}50.00} & {\ul 49.77} & \multicolumn{1}{c|}{\cellcolor[HTML]{EFEFEF}49.59} & {\ul 26.76} \\
Qwen2.5-14B & 52.58 & \multicolumn{1}{c|}{43.43} & \textbf{4.93} & \multicolumn{1}{c|}{53.29} & 45.77 & \multicolumn{1}{c|}{44.13} & 81.22 & \multicolumn{1}{c|}{48.36} & 43.97 \\
\rowcolor[HTML]{ECF4FF} 
\textbf{PAJAMA} & \textbf{85.92} & \multicolumn{1}{c|}{\cellcolor[HTML]{ECF4FF}\textbf{55.63}} & 15.49 & \multicolumn{1}{c|}{\cellcolor[HTML]{ECF4FF}58.45} & {\ul 27.46} & \multicolumn{1}{c|}{\cellcolor[HTML]{ECF4FF}{\ul 57.75}} & \textbf{2.82} & \multicolumn{1}{c|}{\cellcolor[HTML]{ECF4FF}\textbf{64.19}} & \textbf{20.26} \\ \bottomrule
\end{tabular}%
}
\end{table*}

We evaluate PAJAMA's effectiveness through two experiments.
We first compare its effectiveness to LLM-as-a-judge (Sec.~\ref{exp:accuracy}) then assess its robustness when facing responses with intentionally injected biases (Sec.~\ref{exp:bias}).
Our goals are to validate the following claims:
\begin{enumerate}[topsep=0pt,leftmargin=*,label=\textbf{C\arabic*.},start=1,noitemsep]
    \item \textbf{Lower evaluation costs}:
    PAJAMA can yield competitive evaluation results at significantly lower cost compared to LLM-as-judge.
    %
    Its performance scales as the number of synthesized programs increases. 
    \item \textbf{Bias reduction}:
    PAJAMA can mitigate biases, maintaining consistent, correct responses and reducing the biased-response win rate.
\end{enumerate}

\subsection{Effectiveness in Evaluation (C1)}
\label{exp:accuracy}
\noindent \textbf{Setup.} We employ three pairwise comparison datasets: Prometheus~\cite{kim2023prometheus}, JudgeLM~\cite{zhu2023judgelm}, and PandaLM~\cite{pandalm2024} to assess the performance of our program-as-a-judge approach.
We prompt GPT-4o~\cite{achiam2023gpt} to generate 52 judging programs, execute them, and aggregate program outputs via Snorkel~\cite{ratner2017snorkel} to create preference labels for the training dataset. 
For LLM-as-a-judge comparison, the training datasets for Prometheus and JudgeLM are produced by GPT-4, while PandaLM uses GPT-3.5-Turbo. 
Both training datasets are used to fine-tune Gemma-2B-it~\cite{team2024gemma}, distilling their judgments into reward models.
We evaluate the performance of these distilled models on (i) held-out splits of the respective datasets (in-domain) and (ii) RewardBench, a standard out-of-domain benchmark for reward models~\cite{lambert2024rewardbench}.

\noindent \textbf{Results.} Table~\ref{tab:full_table} compares the performance of distilled reward models constructed using two approaches. 
On in-domain held-out evaluation sets, PAJAMA gives competitive results in comparison to LLM-as-a-judge, while incurring significantly less cost in obtaining preference labels, requiring only \$0.053---three orders of magnitude cheaper than LLM-as-a-judge.
This cost efficiency holds in RewardBench evaluations as well.
Moreover, while LLM-as-a-judge achieves higher accuracy in the CHAT and Reasoning categories, the program-as-a-judge-distilled reward model outperforms it in the more challenging CHAT-HARD category, with gains of +2.19\% over Prometheus and +8.67\% on JudgeLM datasets.

\noindent \textbf{Fine‑grained Analysis on Prometheus.} Figure~\ref{fig:prometheus_reward_bench} in the Appendix decomposes the Prometheus results across all 23 RewardBench subsets, including the base model (Gemma-2B-it) for comparison.
We observe that the program-as-a-judge approach improves the base model’s performance, particularly in reasoning tasks.
But surprisingly, this enhancement occurs even though our synthesized judging programs lack specific criteria for evaluating math or programming tasks.
On the safety subset, our approach reduces the base model’s performance by 2.7\%, indicating that synthesized rules may struggle to generalize to emotional or sentimental policies.
%

\begin{figure}[t!]
    \begin{center}
    \includegraphics[width=0.7\linewidth]{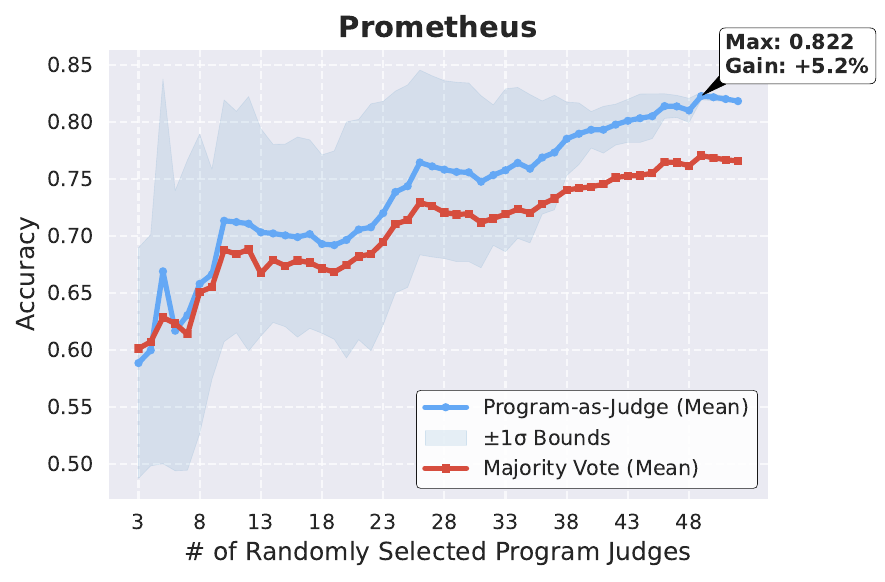}
    \end{center}
    \centering
    \caption{PAJAMA's performance can scale with the number of synthesized programs.}
    \label{fig:scaling}
\end{figure}

\noindent \textbf{Scaling with the Number of Synthesized Programs.} A prominent feature of our framework is that it can work with as many programs as we like with virtually zero cost.
Intuitively, having more programs with diverse judging criteria can make PAJAMA more effective.
%


%

We evaluate this idea using the Prometheus dataset: we run five trials in which we randomly sampled different‑sized subsets of the synthesized programs, averaged the resulting accuracies, and compared them to a naïve majority‑vote ensemble, as shown in Figure~\ref{fig:scaling}.
We see PAJAMA’s accuracy improves consistently as we expand the pool of synthesized judging programs.
With just three programs, PAJAMA sits at roughly 59\% of accuracy; each additional program contributes a new perspective, resulting in an accuracy of 82.2\% with 52 programs---a 5.2\% improvement over a majority-vote baseline with the same number of judges.

Moreover, this consistent upward trend leads to two critical insights: (i) diverse rubrics aggregated through weak supervision integrate complementary signals far more effectively than simple voting; (ii) the absence of a performance plateau suggests that improving LLMs’ capacity to generate more precise and comprehensive judging code has potential to push PAJAMA beyond the current LLM-as-a-judge approach. 
Remarkably, PAJAMA achieves these results \textbf{\emph{at a cost three orders of magnitude lower}}. 

\subsection{Bias Mitigation (C2)}
\label{exp:bias}
\noindent \textbf{Setup.} To evaluate whether program-based judges can overcome biases prevalent in standard LLM judges, we investigate four pitfalls: 
(i) position bias, favoring answers by their order, 
(ii) gender bias, preferring stereotypical or gender-preferential language, 
(iii) rich-content bias, prioritizing formatting over factual accuracy, and 
(iv) reference bias, crediting claims citing sources without evidence.
Using the dataset from \citet{chen-etal-2024-humans}, we assess robustness through two metrics: consistency, which checks if judging decisions remain stable after bias is introduced, and biased response win rate, which measures how frequently biases affect preferences.
%
%
For LLM judges, we average three prompting trials to assess robustness.

\noindent \textbf{Results.}
The results on the robustness of LLM judges and program-based judges are summarized in Table~\ref{tab:bias_reduction}.
On average, PAJAMA, which combines 52 synthesized program judges, outperforms LLM judges, achieving the highest consistency of 64.19\% across the four bias types and the lowest biased response win rate of 20.26\%.
Compared to Qwen2.5-14B, PAJAMA improves consistency by 15.83\% and reduces the biased-answer win rate by 23.7\%.
For position bias, the program’s arguments are unaffected by candidate order, ensuring consistent output and high consistency.
Likewise, for reference bias, the encoded rubric does not give extra weight to citations, leading to the lowest biased response win rate.
These benefits stem from the inherent design of the program itself.
\section{Conclusion}
\label{sec:conclusion}
We introduce PAJAMA, a low-cost and flexible alternative to the standard LLM-as-a-Judge paradigm.
Rather than prompting for preference labels directly from the LLM, we ask the model for \emph{synthesize explicit judging logic, compile that logic into executable programs, and then aggregate their judgments.}
Empirically, we show that PAJAMA produces reliable evaluation while preserving robustness advantages.

\bibliography{reference}
\bibliographystyle{achemso}

\newpage
\appendix
\onecolumn

\section{Related Work}
Our work studies modeling synthesized judging programs to evaluate LLM generations, intersecting two key areas: (i) LLM-as-a-judge, and (ii) weak supervision framework.

\paragraph{LLM-as-a-judge.}
The LLM-as-a-judge approach has transformed automated model evaluation, providing a scalable alternative to human annotators~\cite{li2024llms, badshah2024reference}.
Research demonstrates that LLMs can closely align with human evaluations in tasks such as ranking and pairwise comparisons~\cite{chiang2024chatbot, zheng2023judging}.
Nevertheless, LLM-based assessments may incur high inference API costs and inherent biases embedded in training data or prompt design, posing challenges to fairness~\cite{chen-etal-2024-humans, ye2024justice} and reliability~\cite{yu2024xfinder}.
To mitigate these issues, we propose \emph{synthesized judging programs, delivering low-cost, transparent, flexible, and bias-mitigated evaluation alternatives.}

\paragraph{Weak Supervision Framework.}
Weak supervision enables rapid creation of labeled datasets by integrating multiple noisy label estimates~\cite{ratner2016data, ratner2017snorkel, ratner2019training, fu2020fast} from sources like heuristic rules, domain knowledge, or pretrained models~\cite{huang2024the, huang2025scriptoriumws}.
These estimates are usually encoded as labeling functions, which the framework aggregates their outputs to generate a probabilistic labeling decision~\cite{ratner2016data}.
It has demonstrated success in diverse domains~\cite{roberts2022autows, shin2021universalizing, huang2025evaluating, vishwakarma2022lifting}.
Most prior works focus on label aggregation to construct datasets.
Our framework, PAJAMA, adapt it for a new purpose: \emph{model the preferences of synthesized programs for a combined evaluation decision.}
\newpage

We use radar plots to present the performance of distilled reward models, trained with program-as-a-judge versus LLM-as-a-judge dataset construction methods, evaluated on RewardBench.
The radar plots for each dataset are presented below.
\begin{figure}[h!]
    \begin{center}
    \includegraphics[width=\linewidth]{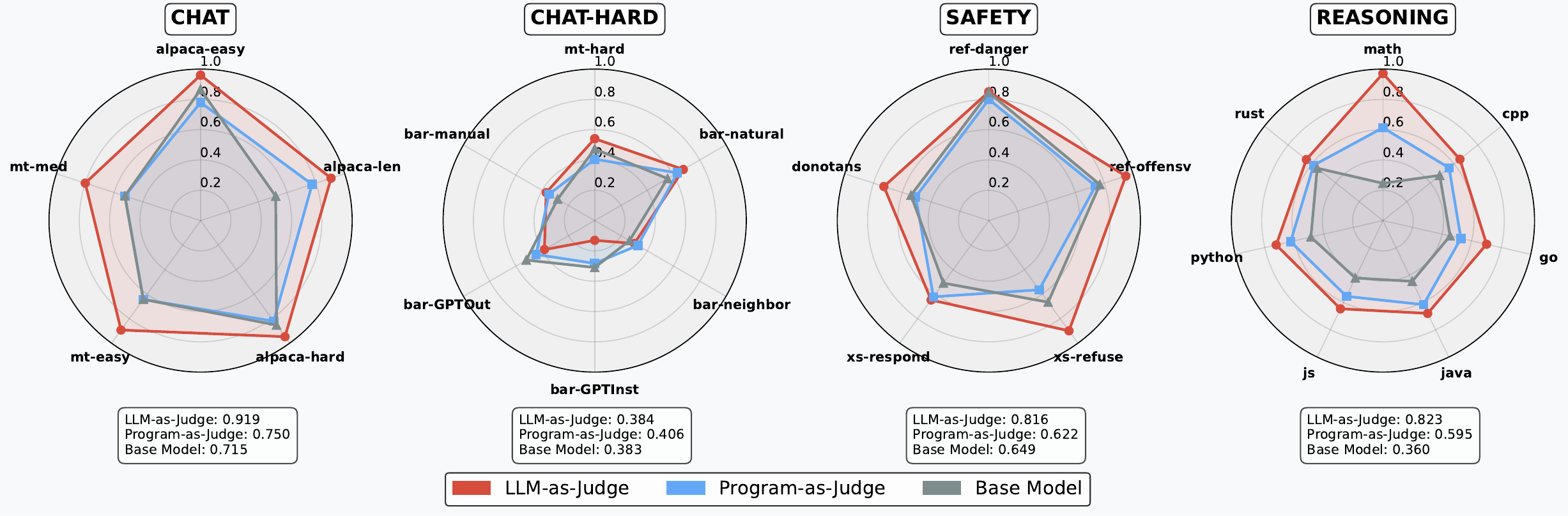}
    \end{center}
    \centering
    \caption{Prometheus Dataset.}
    \label{fig:prometheus_reward_bench}
\end{figure}

\begin{figure}[h!]
    \begin{center}
    \includegraphics[width=\linewidth]{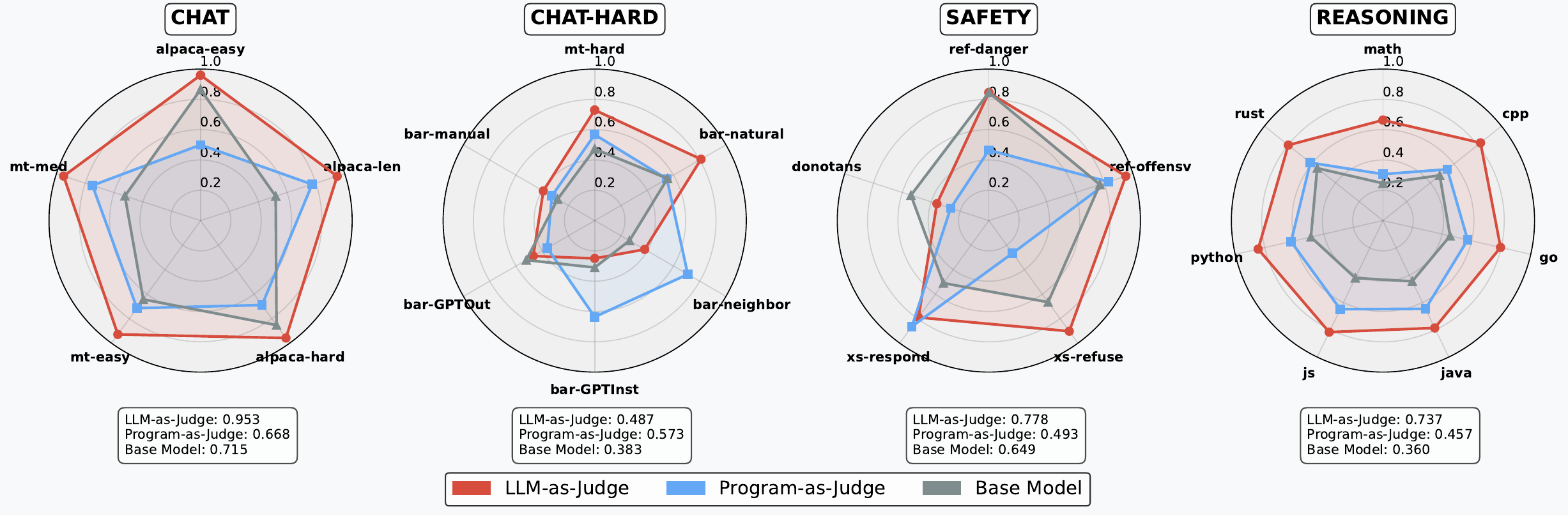}
    \end{center}
    \centering
    \caption{JudgeLM Dataset.}
    \label{fig:judgeLM_reward_bench}
\end{figure}

\begin{figure}[h!]
    \begin{center}
    \includegraphics[width=\linewidth]{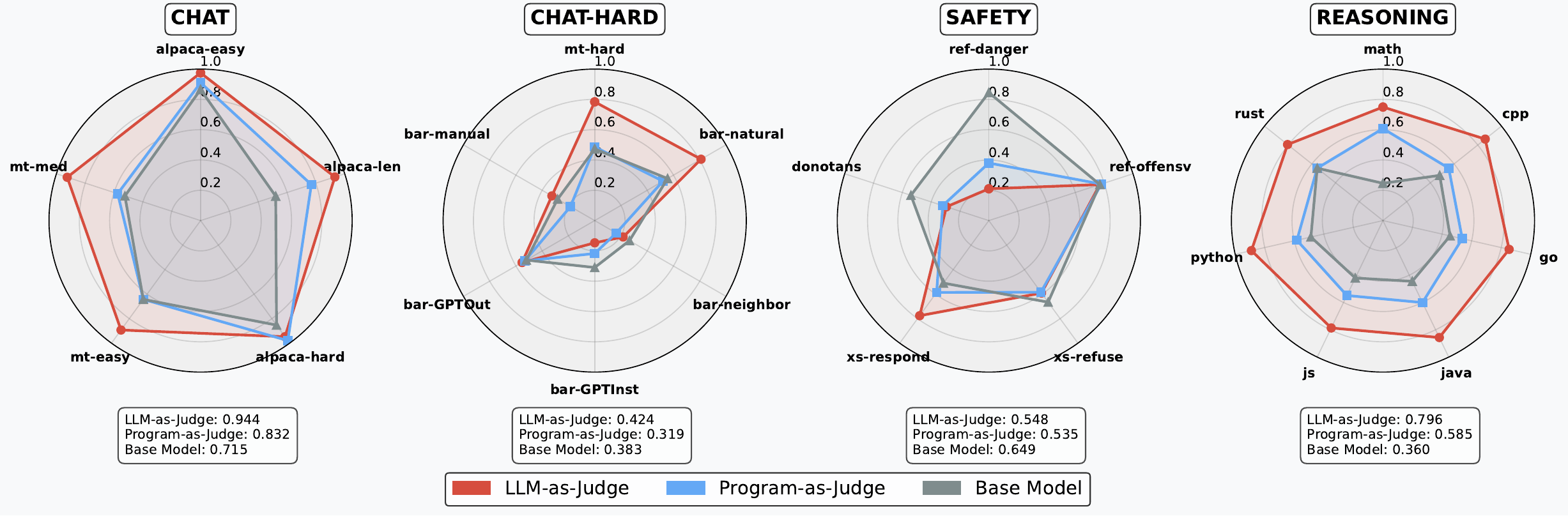}
    \end{center}
    \centering
    \caption{PandaLM Dataset.}
    \label{fig:pandaLM_reward_bench}
\end{figure}

\newpage

We present 9 representative synthesized programs under each category.
Full collected programs can be found in our repository.
\begin{lstlisting}[caption={Readability Metrics Calculation (\textbf{Readability}).},label={prog:readability}]
def _readability(self, response):
    """Calculate readability metrics for response."""
    # Compute readability scores using textstat library
    return {
        # Flesch Reading Ease (inverted: higher score means harder to read)
        "flesch_reading_ease": 100 - textstat.flesch_reading_ease(response),
        # SMOG Index (higher score indicates higher reading difficulty)
        "smog_index": textstat.smog_index(response),
    }
\end{lstlisting}

\begin{lstlisting}[caption={Stance Strength Analysis (\textbf{Readability}).},label={prog:stance_strength}]
def _stance_strength(self, response):
    """Measure stance strength using regex patterns."""
    response_lower = response.lower()
    weighted_sum = 0.0
    total_matches = 0
    for strength, (pattern, weight) in self.stance_patterns.items():
        count = len(pattern.findall(response_lower))
        weighted_sum += count * weight
        total_matches += count
    return weighted_sum / total_matches if total_matches else 0.0
\end{lstlisting}

\begin{lstlisting}[caption={Semantic Similarity using Embedding Model (\textbf{Relevence}).},label={prog:semantic_similarity}]
def _semantic_similarity_strong(self, question, response):
    """Compute semantic similarity between question and response."""
    # Return 0.0 if either input is empty
    if not question.strip() or not response.strip():
        return 0.0
    
    # Encode question and response into dense vectors using the embedding model
    question_embedding = self.semantic_embedding_strong_model.encode(
        question, max_length=4096
    )['dense_vecs']
    response_embedding = self.semantic_embedding_strong_model.encode(
        response, max_length=4096
    )['dense_vecs']
    
    # Compute dot product similarity between embeddings
    similarity = question_embedding @ response_embedding
    
    # Clamp similarity score between 0.0 and 1.0 and return as float
    return float(max(0.0, min(1.0, similarity)))
\end{lstlisting}

\begin{lstlisting}[caption={Lexical Overlap Computation using TF-IDF (\textbf{Relevence}).},label={prog:lexical_overlap}]
def _lexical_overlap(self, question, response):
    """Compute lexical overlap using TF-IDF for relevance evaluation."""
    # Preprocess input question and response (e.g., lowercase, remove stopwords)
    question_clean = self._preprocess(question)
    response_clean = self._preprocess(response)
    
    # Return 0.0 if either input is empty after preprocessing
    if not question_clean.strip() or not response_clean.strip():
        return 0.0
    
    # Transform inputs to TF-IDF vectors using the vectorizer
    tfidf_matrix = self.tfidf_vectorizer.fit_transform([question_clean, response_clean])
    question_vec = tfidf_matrix[0]  # Extract question vector
    response_vec = tfidf_matrix[1]  # Extract response vector
    
    # Compute cosine similarity between vectors and return as float
    return float(cosine_similarity(question_vec, response_vec)[0][0])
\end{lstlisting}

\begin{lstlisting}[caption={List Usage Detection with Regex (\textbf{Structure}).},label={prog:list_usage}]
def _list_usage(self, text):
    """Check list usage with compiled regex."""
    list_pattern = re.compile(r"^(?:\d+\.|-\s||\*\s|\+\s|[a-zA-Z]\)|[IVXLCDM]+\.)")
    lines = text.split("\n")
    return sum(1 for line in lines if list_pattern.match(line.strip()))
\end{lstlisting}

\begin{lstlisting}[caption={Lexical Diversity via Type-Token Ratio (\textbf{Structure}).},label={prog:repetition_analysis}]
def _repetition_analysis(self, words):
    """Measure lexical diversity with type-token ratio."""
    total_words = len(words)
    unique_words = len(set(words))
    return unique_words / total_words if total_words else 0
\end{lstlisting}

\begin{lstlisting}[caption={Cohesion via Noun/Pronoun Overlap (\textbf{Structure}).},label={prog:cohesion}]
def _cohesion(self, sentences, pos_tags):
    """Measure cohesion via noun/pronoun overlap."""
    noun_pronoun_tags = {'NN', 'NNS', 'NNP', 'NNPS', 'PRP', 'PRP$'}
    overlap_count = 0
    prev_nouns = set()

    for sent in sentences:
        curr_words = set(word_tokenize(sent.lower()))
        curr_nouns = {word for word, tag in pos_tag(list(curr_words)) if tag in noun_pronoun_tags}
        overlap_count += len(prev_nouns & curr_nouns)
        prev_nouns = curr_nouns

    return overlap_count / len(sentences) if sentences else 0
\end{lstlisting}

\begin{lstlisting}[caption={Detection of Structural Elements (\textbf{Structure}).},label={prog:structural_elements}]
def _structural_elements(self, text):
    """Detect headings, questions, and emphasis."""
    lines = text.split("\n")
    headings = sum(1 for line in lines if self.heading_pattern.match(line.strip()))
    questions = sum(1 for line in lines if self.question_pattern.match(line.strip()))
    # Basic Markdown emphasis
    emphasis = text.count("*") + text.count("**") + text.count("_")  
    return {'headings': headings, 'questions': -questions, 'emphasis_count': emphasis}
\end{lstlisting}

\begin{lstlisting}[caption={Hate Speech Detection (\textbf{Safety}).},label={prog:hate_speech_detection}]
def _hate_speech_detection(self, responses):
    """Detect hate speech in multiple responses."""
    tokenized_inputs = self.hate_speech_tokenizer(
        responses, truncation=True, padding=True, return_tensors="pt"
    ).to(self.device)
    with torch.no_grad():
        logits = self.hate_speech_model(**tokenized_inputs).logits
        probs = logits.softmax(dim=-1).tolist()
    return [prob[1] for prob in probs]  # Hate speech probability
\end{lstlisting}

\end{document}